\documentclass[sigconf]{acmart}
\setkeys{acmart.cls}{balance=false}

\AtBeginDocument{%
  \providecommand\BibTeX{{%
    Bib\TeX}}}

\setcopyright{cc}
 \setcctype{by}

 \acmConference[BCB '26]{17th ACM International Conference on Bioinformatics, Computational Biology and Health Informatics}{June 30-July 03, 2026}{Rende (CS), Italy}

\acmDOI{}
\acmISBN{}
	
\usepackage{acronym}
\usepackage{xspace}
\usepackage{comment}
\usepackage{booktabs}

\usepackage{multirow}
\usepackage{makecell}
\usepackage{tablefootnote}

\def\BibTeX{{\rm B\kern-.05em{\sc i\kern-.025em b}\kern-.08em
    T\kern-.1667em\lower.7ex\hbox{E}\kern-.125emX}}

\acrodef{LSTM}{Long Short-Term Memory}
\acrodef{TCN}{Temporal Convolutional Networks}
\acrodef{EDA}{electrodermal activity}
\acrodef{ECG}{electrocardiography}
\acrodef{BVP}{blood volume pulse}
\acrodef{TEMP}{body temperature}
\acrodef{ACC}{accelerometer}
\acrodef{RES}{respiration}
\acrodef{RNNs}{Recurrent Neural Networks}

\begin{document}

\title{Deep Temporal Modeling and Ensemble Fusion for Multimodal Emotion Recognition from Physiological Signals}

\author{Desta Haileselassie Hagos}
\affiliation{%
 \institution{Howard University}
 \city{Washington}
 \state{DC}
 \country{USA}}
 \email{desta.hagos@howard.edu}
 \orcid{0000-0002-2014-2749}

\author{Saurav Keshari Aryal}
\affiliation{%
  \institution{Howard University}
  \city{Washington}
  \state{DC}
  \country{USA}}
  \email{saurav.aryal@howard.edu}
  \orcid{0000-0001-9815-9295}

\author{Patrick Ymele-Leki}
\affiliation{%
  \institution{Howard University}
  \city{Washington}
  \state{DC}
  \country{USA}}
\email{patrick.ymeleleki@howard.edu}
\orcid{0000-0002-8504-0540}

\author{Anietie Andy}
\affiliation{%
  \institution{Howard University}
  \city{Washington}
  \state{DC}
  \country{USA}}
\email{anietie.andy@howard.edu}
\orcid{0000-0002-7043-3042}

\author{Legand L. Burge}
\affiliation{%
  \institution{Howard University}
  \city{Washington}
  \state{DC}
  \country{USA}}
\email{lburge@howard.edu}
\orcid{0009-0000-9876-7135}

\renewcommand{\shortauthors}{}

\begin{abstract}
  Physiological stress and emotion recognition are important for health monitoring and affective computing. In this work, we present a comprehensive evaluation of deep learning models such as \ac{LSTM}, \ac{TCN}, and Transformer on the WESAD dataset for multimodal affect recognition using wrist and chest sensor signals. We perform ablation studies to assess the individual contributions of each modality by training models on wrist-only and chest-only inputs. In addition, we implement a late-fusion ensemble strategy that combines predictions from all three architectures trained on multimodal input. We also employ early fusion at the sensor level by concatenating wrist and chest signals before feeding them into each model. Our results show that Transformer models consistently achieve the highest accuracy in multimodal settings, while TCN models perform best in the wrist-only configuration. The ensemble method yields the highest overall accuracy (98.91 $\pm$ 0.13\%) and macro-F1 score (98.56 $\pm$ 0.17\%). These findings demonstrate the effectiveness of sensor fusion and ensemble-based fusion in developing robust systems for physiological emotion recognition.

  \vspace{1.5ex}

  \noindent \textbf{Note}: \textit{An extended version containing supplementary analyses is included in the appendices.}
  
\end{abstract}



\maketitle

\fancyhead{}
\fancyfoot{}
\pagestyle{plain}

\begin{center}
\textit{
Accepted for publication in the 17th ACM International Conference on
Bioinformatics, Computational Biology and Health Informatics
(ACM BCB 2026).
}

\smallskip

DOI: \url{https://doi.org/10.1145/3807503.3819363}

\end{center}

\section{Introduction}
\label{introduction}

The ability to detect and classify human emotional states from physiological signals is increasingly important across domains including mental health monitoring, wearable computing, human-computer interaction, and affective computing~\cite{calvo2010affect}. Stress is a growing public health concern associated with chronic diseases, reduced quality of life, and decreased productivity~\cite{schneiderman2005stress}. The accurate and real-time detection of stress and affective states can pave the way for timely interventions and personalized wellness solutions~\cite{healey2005detecting}. Traditional methods for detecting affects are based heavily on self-reports or interviews, which are subjective and impractical for continuous monitoring~\cite{schmidt2019wearable, sheikh2021wearable}. Recent advances in wearable sensors have allowed the collection of rich physiological signals, such as \ac{EDA}, respiration, body temperature, heart rate, and accelerometry data~\cite{schmidt2018introducing}. These signals enable real-time inference of emotional states using data-driven models~\cite{rissler2018got, larradet2020toward}. Deep learning models are particularly effective for modeling time-series data. Architectures such as \ac{LSTM}~\cite{hochreiter1997long}, \ac{TCN}~\cite{lea2017temporal, bai2018empirical}, and Transformers~\cite{vaswani2017attention} are particularly well-suited to learning complex temporal dependencies in multimodal physiological data. However, the effective use of different sensor modalities, understanding their relative contributions, and optimizing their fusion remain open challenges in wearable affect recognition~\cite{kulvicius2025deep, li2024review}. 

The goal of this work is to systematically evaluate temporal deep learning architectures (\ac{LSTM}, \ac{TCN}, Transformer) under unimodal, multimodal, and ensemble fusion settings to identify robust and generalizable approaches for physiological emotion recognition. We benchmark these architectures on the WESAD dataset, a multimodal benchmark for wearable stress and affect detection, performing extensive ablation studies using wrist-only and chest-only modalities to understand their standalone performance and how each model adapts to unimodal input. Finally, we propose an ensemble fusion approach that integrates predictions from all three models to improve classification robustness and accuracy.

Our findings provide practical insight into the effectiveness of each model and modality configuration, offering guidance for the development of efficient, reliable, and scalable emotion recognition systems based on wearable sensor data. To our knowledge, no previous study provides a unified and controlled comparison of \ac{LSTM}, \ac{TCN}, and Transformer architectures across unimodal, multimodal, and ensemble settings under a leave-one-subject-out cross-validation (LOSO-CV) protocol on WESAD~\cite{schmidt2018introducing}. Existing works typically examine a single architecture, a single sensor location, or a single fusion strategy in isolation. Our contribution lies in establishing a systematic and architecture-agnostic benchmarking framework that clarifies when and why specific temporal models or sensor modalities are advantageous, effectively isolating the effects of architecture choice, sensor modality, and fusion design on affect recognition performance.

\section{Motivation}
\label{motivation}

Stress and affective disorders are among the leading contributors to the global burden of diseases, affecting both mental and physical health across populations~\cite{calvo2010affect, schneiderman2005stress}. Early and accurate detection of emotional states, such as stress and amusement, can be important in preventing burnout, improving productivity, and enabling timely interventions in clinical and workplace settings~\cite{schneiderman2005stress, sonnentag2015recovery}. Wearable devices enable continuous and non-invasive monitoring of physiological signals. The increasing availability of multimodal sensors embedded in smartwatches, chest straps, and fitness trackers demands intelligent systems capable of interpreting these signals as meaningful emotional labels~\cite{healey2005detecting, schmidt2019wearable}. However, developing such systems involves challenges in modeling temporal dependencies, handling heterogeneous signal modalities, and ensuring generalization across users and contexts~\cite{schmidt2018introducing}. Previous studies have applied LSTMs~\cite{rostami2024lstm, malviya2023mental, zitouni2022lstm}, TCNs~\cite{ding2024masa, ingolfsson2021ecg, alghoul2025enhancing}, and Transformers~\cite{vaswani2017attention, li2025transformer, wu2023transformer, vazquez2022emotion} to physiological affect recognition, but few have systematically compared these architectures under controlled and consistent conditions. Recent work by Liao et al.~\cite{liao2025emotion} and Choi~\cite{choi2025emotion} explore ensemble and fusion strategies but focus on specific model designs rather than a unified cross-architecture evaluation. We adopt WESAD~\cite{schmidt2018introducing} as our benchmark as it enables a direct and reproducible comparison with previous work under a consistent LOSO-CV protocol.

\section{Dataset and Methodology}
\label{approaches}

We adopt a strategy that combines advanced temporal deep learning architectures with both unimodal and multimodal data perspectives to develop a robust system for affect recognition using physiological signals. Our pipeline consists of three main stages: (1) data preparation and segmentation, (2) architecture design with ablation-based evaluation, and (3) multimodal ensemble fusion. 

\subsection{Dataset Description}

We use the publicly available WESAD dataset~\cite{schmidt2018introducing}, a widely adopted benchmark for physiological stress and affect recognition using wearable sensors. The dataset contains multimodal physiological recordings collected from 15 participants (12 male, 3 female) during controlled experimental conditions designed to induce baseline, stress, and amusement states. Physiological signals are acquired using two wearable devices: the Empatica E4 wristband and the RespiBAN chest sensor. The wrist modality includes \ac{EDA}, \ac{BVP}, \ac{TEMP}, and \ac{ACC} signals, while the chest modality includes \ac{RES}, \ac{ECG}, and \ac{ACC} signals. A summary of the physiological signals and their corresponding devices is provided in Table~\ref{tab:wesad_signals}. Each participant undergoes three affective conditions: baseline (neutral state), stress (induced by the Trier Social Stress Test), and amusement (elicited through video stimuli). These conditions serve as ground-truth labels for supervised learning, where the task is to classify each time segment into one of the three affective states. To evaluate generalization across individuals, we adopt a leave-one-subject-out cross-validation (LOSO-CV) protocol, where models are trained on data from all but one participant and tested on the held-out participant. This process is repeated for all participants to ensure subject-independent evaluation. The full dataset contains over 5.3 million samples across all modalities. As the dataset exhibits moderate class imbalance among the three affective states, we employ stratified batching during training and report class-wise performance metrics to ensure fair and reliable evaluation.

\begin{table}[h]
\centering

\caption{Physiological signals collected from wrist (Empatica E4) and chest (RespiBAN) devices in the WESAD dataset.}
\resizebox{\columnwidth}{!}{
\label{tab:wesad_signals}
\begin{tabular}{lll}
\toprule
Device & Signals & Sampling Rate \\
\midrule
Empatica E4 (Wrist) & EDA, BVP, TEMP, ACC & 4--64 Hz \\
RespiBAN (Chest) & RES, ECG, ACC & 700 Hz \\
\bottomrule
\end{tabular}
}

\end{table}

\subsection{Data Preparation}

To prepare the data, we resampled all signals to a common rate of 4 Hz, temporally aligning wrist signals (\ac{EDA}, \ac{TEMP}, \ac{ACC}, \ac{BVP}) and chest signals (\ac{RES}, \ac{ECG}, \ac{ACC}) to a consistent temporal resolution before segmentation. This resampling focuses the models on low-frequency temporal dynamics associated with affective states, as commonly adopted in affective computing studies~\cite{tanwar2024attention}. 

The continuous signals were then segmented into non-overlapping windows of 10 time steps, where each time step corresponds to one preprocessed multimodal sample at 4 Hz. This window length balances temporal resolution and computational efficiency for affective state classification. A sensitivity analysis evaluating the effect of resampling rate on model performance is provided in the supplementary material. Categorical labels were one-hot encoded, and stratified sampling was applied within the LOSO-CV folds to ensure class balance and reproducibility during training. Window segmentation and normalization were performed independently within each LOSO fold to avoid information leakage between training and testing subjects. To ensure a fair comparison across architectures and modality configurations, all models used the same window length, normalization procedure, and segmentation strategy, so that differences in performance reflect architectural design rather than preprocessing choices.

\subsection{Model architectures}

We adopt three model architectures known for their effectiveness in time-series learning. Each model is trained in unimodal and multimodal settings. 

\vspace{1.0ex}
\noindent \textbf{LSTM}. These networks are a class of \ac{RNNs} designed to model long-range temporal dependencies in sequences through memory cells and gating mechanisms~\cite{hochreiter1997long}. They have shown strong performance in physiological signal modeling, particularly for emotion and stress recognition. LSTMs are well-suited for capturing the temporal continuity and gradual transitions characteristic of affective states, including stress onset and relaxation periods. We implement a two-layer bi-directional LSTM followed by a fully connected output layer with dropout regularization to mitigate overfitting. The bi-directional structure enables the model to leverage both past and future context, enhancing its ability to learn complex temporal patterns in physiological signals. Despite the added depth, the model remains relatively lightweight, making it suitable for wearable applications.

\vspace{1.0ex}
\noindent \textbf{TCN}. These models are fully convolutional architectures designed for sequence modeling using dilated causal convolutions, allowing the model to efficiently capture long-range temporal dependencies~\cite{lea2017temporal, bai2018empirical}. Unlike LSTMs, TCNs support parallel processing in time steps, enabling faster training and improved stability. Their ability to capture multi-scale temporal features makes them particularly well-suited for modeling physiological signals of varying durations. We implemented a stack of 1D convolutional layers with progressively increasing dilation rates, combined with residual connections and batch normalization to improve training stability and enhance generalization.

\vspace{1.0ex}
\noindent \textbf{Transformer}. Transformers rely on self-attention mechanisms rather than recurrence or convolution, allowing them to model dependencies over arbitrary time steps without assuming local structure~\cite{vaswani2017attention}. Transformers identify salient signal patterns regardless of position. In our implementation, we use a compact encoder with three layers of multi-head self-attention, learnable positional embeddings, and dropout regularization. The model concludes with global average pooling across the temporal dimension to aggregate sequence-level representations, followed by a softmax classification layer. This architecture is particularly well-suited for multimodal emotion recognition, as its self-attention mechanism can capture global dependencies across asynchronous signal streams and dynamically focus on the most informative temporal segments, an essential capability when fusing heterogeneous biosignals.

\vspace{1.0ex}

\noindent \textbf{Fusion Strategies}. We adopt two complementary fusion strategies. First, for multimodal input-level fusion (early fusion), we concatenate wrist- and chest-derived features along the channel dimension before feeding them into the model. This enables the temporal architectures (LSTM, Transformer, TCN) to learn joint representations across sensor types. Second, for output-level fusion (late fusion), we perform ensemble classification by averaging the \emph{softmax} probability outputs of three independently trained models. This dual-fusion design enables the system to benefit from both the complementary nature of multimodal physiological signals and the diverse learning capacities of the architectures, which in turn reduces overfitting and enhances generalization in affect recognition tasks. Throughout this paper, late fusion refers to the ensemble-learning approach in which model-level \emph{softmax} probabilities are averaged to produce the final prediction. This is distinct from the traditional multimodal-learning definition of late fusion, where separate models operate on different modalities. In our ensemble, all three architectures receive the same multimodal input.

\vspace{-1.0ex}

\subsection{Implementation Details}

All models were implemented in \texttt{Python} using \texttt{TensorFlow 2.16.2} and trained on two NVIDIA GeForce RTX 4090 GPUs (24 GB GDDR6X each) with fixed random seeds to ensure reproducibility. Architectural configurations were selected based on empirical tuning guided by validation performance and previous work in physiological signal processing, exploring variations in layer depth, hidden units, kernel sizes, and regularization strength for each model family. We report the mean and standard deviation of all evaluation metrics across the 15 LOSO folds to provide a reliable estimate of model performance and inter-subject variability. Training was performed using the Adam optimizer with early stopping based on validation loss to prevent overfitting. Input features were standardized on a per-subject basis within each LOSO fold to prevent data leakage, and categorical labels were one-hot encoded. For multimodal configurations, wrist- and chest-derived features were concatenated at the feature level (early fusion), while ensemble fusion was performed via averaging of \emph{softmax} probabilities across independently trained models (late fusion).

\vspace{1.0ex}

\noindent\textbf{Hyperparameters.} The Transformer used embedding dimension 64, 4 attention heads, 3 encoder blocks, and dropout 0.3. The TCN used 128 filters, kernel size 7, and dilation factors (1,2,4,8). The LSTM used two stacked bidirectional layers with 64 units each. All models used Adam (learning rate = 0.001), early stopping (patience=3), and \texttt{ReduceLROnPlateau} (factor=0.5, patience=2). Batch size was 32 for Transformer and 64 for LSTM and TCN.

\section{Experimental Results and Analysis}
\label{discussion_of_experimental_results}

We present a comprehensive evaluation of the proposed models in the WESAD dataset~\cite{schmidt2018introducing} to assess their effectiveness in recognizing affective states from physiological signals. Our experiments compare multiple model architectures, LSTM, Transformer, and TCN, in both unimodal and multimodal configurations under a LOSO-CV protocol. In addition, we conducted ablation studies to analyze the contribution of individual sensor modalities and examine an ensemble strategy that integrates complementary strengths across models. These results provide insights into the impact of architectural choices and modality fusion on performance, generalization, and the design of robust affective computing systems.

\subsection{Ensemble Fusion Performance}

We applied a late fusion ensemble strategy integrating predictions from LSTM, TCN, and Transformer models, each trained independently on the same multimodal input. Instead of concatenating features, the fusion is performed by averaging the \emph{softmax} probability outputs of the three models. This approach leverages the complementary strengths of the architectures: LSTM captures long-term dependencies; TCN models local and mid-range patterns using dilated convolutions; and the Transformer is effective at identifying salient temporal segments through attention mechanisms. As shown in Table~\ref{tab:ensemble_fusion}, the ensemble achieved the highest performance on all metrics, with 98.91 $\pm$ 0.13\% accuracy, 98.56 $\pm$ 0.17\% macro-F1, and 98.90 $\pm$ 0.13\% weighted-F1. This performance gain reflects the combined effect of variance reduction through ensemble averaging and the complementary inductive biases of the three architectures. While variance reduction alone contributes meaningfully to ensemble gains, the diversity of temporal modeling strategies, namely sequential modeling in LSTM, local pattern extraction in TCN, and global attention in Transformer, provides complementary representations that contribute to the stability and robustness of the ensemble across subjects. These results demonstrate the benefit of combining early sensor fusion with late model ensemble strategies, as the approach leverages both cross-sensor synergy and variance reduction to improve performance.

The ensemble provides superior macro-F1 performance and markedly lower variance across LOSO folds, with a standard deviation of $\pm$0.13\% in accuracy and $\pm$0.17\% in macro-F1, compared to $\pm$0.51\%--0.59\% and $\pm$0.64\%--0.76\% for individual models, respectively, demonstrating stable and consistent generalization across subjects. Because macro-F1 is more sensitive to class imbalance and reflects consistency across subjects, we report the ensemble as the best overall model, while noting that the Transformer remains the strongest standalone architecture.

\begin{table}[ht]
\caption{Ensemble fusion results combining LSTM, TCN, and Transformer predictions on the multimodal WESAD dataset. All metrics are reported in percentage (\%).}
\resizebox{\columnwidth}{!}{
\begin{tabular}{l|ccc|c}
\toprule
\textbf{Class} & \textbf{Precision} & \textbf{Recall} & \textbf{F1-score} & \textbf{Accuracy} \\
\midrule
Baseline        & 98.87 & 99.36 & 99.12 & \multirow{5}{*}{98.91 $\pm$ 0.13} \\
Stress          & 99.50 & 99.75 & 99.62 &  \\
Amusement       & 97.94 & 95.96 & 96.94 &  \\
\cmidrule(lr){1-4}
\emph{Macro Avg}       & 98.77 & 98.36 & 98.56 $\pm$ 0.17&  \\
\emph{Weighted Avg}    & 98.90 & 98.91 & 98.90 $\pm$ 0.13&  \\
\bottomrule
\end{tabular}
}
\label{tab:ensemble_fusion}
\end{table}

\subsection{Model Architecture Comparison}

The performance of the model varied across the modalities and configurations. In the multimodal setting (Table~\ref{tab:multimodal_performance}), Transformer models achieved the highest overall performance, likely due to their ability to model temporal dependencies through self-attention mechanisms. In modality-constrained scenarios, the ablation study (Table~\ref{tab:ablation_wrist_chest}) shows that TCN performed best in the wrist-only setting, while LSTM achieved the strongest results for chest-only inputs. Table~\ref{tab:unimodal_comparison} indicates that LSTM achieved the highest accuracy and F1-scores in the unimodal wrist-only configuration, followed by Transformer and TCN. These findings suggest that architectural performance is protocol-specific and reflects the interaction between input representation and model configuration. Within the ablation protocol, TCN retained the strongest wrist-only performance, while within the primary unimodal protocol, the bidirectional LSTM performed best. These are setting-specific observations and should not be interpreted as a universal architectural ranking across protocols.

Table~\ref{tab:unimodal_comparison} reports results from our primary unimodal pipeline, where wrist-only inputs are taken from a dedicated preprocessed wrist representation pipeline and each architecture uses its primary configuration, including a bidirectional stacked LSTM, the main TCN settings, and the main Transformer configuration. Table~\ref{tab:ablation_wrist_chest}, by contrast, reconstructs wrist-only inputs by slicing wrist features from the fused multimodal tensor representation and rebuilding sequences from that sliced representation, resulting in a different effective input distribution. The ablation models also use architecture-specific configurations that differ from the primary benchmark, including differences in LSTM depth and directionality, TCN kernel size, dilation pattern, and filter count, as well as batch size, early stopping schedule, and learning rate scheduling. Within each table, all model comparisons are conducted under identical conditions, which is the basis for our architectural conclusions. These findings suggest that, while attention-based architectures are particularly advantageous in fused or complex input settings, convolutional and recurrent models remain more effective in low-signal or sensor-limited scenarios. Multimodal fusion consistently outperforms unimodal configurations across all architectures, with wrist-only models offering competitive performance in resource-constrained settings (Table~\ref{tab:unimodal_comparison} vs. Table~\ref{tab:ablation_wrist_chest}).

\begin{table}[ht]
\caption{Unimodal classification performance on wrist signals. All evaluation metrics are reported in percentage (\%).}

\centering
\resizebox{\columnwidth}{!}{
\begin{tabular}{l|l|ccc|c}
\toprule
\textbf{Model} & \textbf{Class} & \textbf{Precision} & \textbf{Recall} & \textbf{F1-Score} & \textbf{Accuracy} \\
\midrule
\multirow{6}{*}{LSTM} 
  & Baseline      & 95.92 & 96.66 & 96.29 & \multirow{6}{*}{95.81 $\pm$ 0.48} \\
  & Stress        & 97.03 & 98.37 & 97.70 & \\
  & Amusement     & 93.14 & 88.54 & 90.78 & \\
  \cmidrule(lr){2-5}
  & \emph{Macro Avg}     & 95.36 & 94.52 & 94.92 $\pm$ 0.53& \\
  & \emph{Weighted Avg}  & 95.79 & 95.81 & 95.79 $\pm$ 0.46& \\
\midrule
\multirow{6}{*}{Transformer} 
  & Baseline      & 93.53 & 94.46 & 94.00 & \multirow{6}{*}{93.33 $\pm$ 1.68} \\
  & Stress        & 93.79 & 98.62 & 96.15 & \\
  & Amusement     & 91.56 & 80.27 & 85.54 & \\
  \cmidrule(lr){2-5}
  & \emph{Macro Avg}     & 92.96 & 91.12 & 91.87 $\pm$ 1.93& \\
  & \emph{Weighted Avg}  & 93.28 & 93.33 & 93.22 $\pm$ 1.70 & \\
\midrule
\multirow{6}{*}{TCN} 
  & Baseline      & 88.90 & 95.46 & 92.06 & \multirow{6}{*}{91.06 $\pm$ 0.46} \\
  & Stress        & 94.84 & 92.22 & 93.51 & \\
  & Amusement     & 92.03 & 75.11 & 82.72 & \\
  \cmidrule(lr){2-5}
  & \emph{Macro Avg}     & 91.92 & 87.60 & 89.43 $\pm$ 0.64& \\
  & \emph{Weighted Avg}  & 91.21 & 91.06 & 90.92 $\pm$ 0.48& \\
\bottomrule
\end{tabular}
}
\label{tab:unimodal_comparison}
\end{table}


\begin{table}[ht]
\caption{Multimodal fusion classification performance. All evaluation metrics are reported in percentage (\%).}

\resizebox{\columnwidth}{!}{
\begin{tabular}{l|l|ccc|c}
\toprule
\textbf{Model} & \textbf{Class} & \textbf{Precision} & \textbf{Recall} & \textbf{F1-score} & \textbf{Accuracy} \\
\midrule

\multirow{6}{*}{LSTM} 
  & Baseline      & 96.62 & 99.29 & 97.93 & \multirow{6}{*}{97.70 $\pm$ 0.55} \\
  & Stress        & 99.37 & 99.00 & 99.18 & \\
  & Amusement     & 98.29 & 90.34 & 94.15 & \\
  \cmidrule(lr){2-5}
  & \emph{Macro Avg}     & 98.09 & 96.21 & 97.09 $\pm$ 0.72& \\
  & \emph{Weighted Avg}  & 97.73 & 97.70 & 97.67 $\pm$ 0.56& \\

\midrule

\multirow{6}{*}{Transformer} 
  & Baseline      & 99.29 & 98.94 & 99.11 & \multirow{6}{*}{99.02 $\pm$ 0.51} \\
  & Stress        & 99.00 & 99.62 & 99.31 & \\
  & Amusement     & 98.21 & 98.21 & 98.21 & \\
  \cmidrule(lr){2-5}
  & \emph{Macro Avg}     & 98.83 & 98.92 & 98.88 $\pm$ 0.64 & \\
  & \emph{Weighted Avg}  & 99.02 & 99.02 & 99.02 $\pm$ 0.51& \\

\midrule

\multirow{6}{*}{TCN} 
  & Baseline      & 97.44 & 97.09 & 97.26 & \multirow{6}{*}{96.76 $\pm$ 0.59} \\
  & Stress        & 97.16 & 98.50 & 97.82 & \\
  & Amusement     & 93.85 & 92.58 & 93.21 & \\
  \cmidrule(lr){2-5}
  & \emph{Macro Avg}     & 96.15 & 96.06 & 96.10 $\pm$ 0.76& \\
  & \emph{Weighted Avg}  & 96.75 & 96.76 & 96.75 $\pm$ 0.60& \\

\bottomrule
\end{tabular}
}
\label{tab:multimodal_performance}
\end{table}

\begin{table}[ht]
\caption{Ablation study results for wrist-only and chest-only inputs. All evaluation metrics are reported in percentage (\%).}
\resizebox{\columnwidth}{!}{
\begin{tabular}{l|l|rrr|c}
\toprule
\textbf{Input} & \textbf{Model} & \textbf{Precision} & \textbf{Recall} & \textbf{F1-score} & \textbf{Accuracy} \\
\midrule
\midrule
\multicolumn{6}{c}{\textbf{Wrist-Only}} \\
\midrule
\multirow{6}{*}{LSTM}
  & Baseline      &  96.41 &  93.47 &  94.92 & \multirow{6}{*}{\centering{94.38 $\pm$ 0.88}} \\
  & Stress        &  97.63 &  97.99 &  97.81 & \\
  & Amusement     &  83.33 &  90.81 &  86.91 & \\
  \cmidrule(lr){2-5}
  & \emph{Macro Avg}     &  92.46 &  94.09 &  93.21 $\pm$ 0.97& \\
  & \emph{Weighted Avg}  &  94.58 &  94.38 &  94.44 $\pm$ 0.85& \\
  \midrule

\multirow{6}{*}{Transformer}
  & Baseline      &  88.00 &  92.12 &  90.01 & \multirow{6}{*}{\centering{88.73 $\pm$ 1.79}} \\
  & Stress        &  91.60 &  95.73 &  93.62 & \\
  & Amusement     &  84.88 &  65.47 &  73.92 & \\
  \cmidrule(lr){2-5}
  & \emph{Macro Avg}     &  88.16 &  84.44 &  85.85 $\pm$ 2.46& \\
  & \emph{Weighted Avg}  &  88.56 &  88.73 &  88.39 $\pm$ 1.90& \\
  \midrule

\multirow{6}{*}{TCN}
  & Baseline      &  98.72 &  98.30 &  98.51 & \multirow{6}{*}{\centering{98.20 $\pm$ 0.41}} \\
  & Stress        &  98.88 &  99.37 &  99.13 & \\
  & Amusement     &  95.30 &  95.73 &  95.51 & \\
  \cmidrule(lr){2-5}
  & \emph{Macro Avg}     &  97.63 &  97.80 &  97.72 $\pm$ 0.52& \\
  & \emph{Weighted Avg}  &  98.20 &  98.20 &  98.20 $\pm$ 0.41& \\
\midrule
\multicolumn{6}{c}{\textbf{Chest-Only}} \\
\midrule
\multirow{6}{*}{LSTM}
  & Baseline      &  93.40 &  89.43 &  91.37 & \multirow{6}{*}{\centering{87.93} $\pm$ 0.70} \\
  & Stress        &  85.68 &  89.34 &  87.47 & \\
  & Amusement     &  76.27 &  80.72 &  78.43 & \\
  \cmidrule(lr){2-5}
  & \emph{Macro Avg}     &  85.12 &  86.50 &  85.76 $\pm$ 1.04& \\
  & \emph{Weighted Avg}  &  88.20 &  87.93 &  88.02 $\pm$ 0.74& \\
  \midrule

\multirow{6}{*}{Transformer}
  & Baseline      &  88.48 &  89.43 &  88.95 & \multirow{6}{*}{\centering{84.80 $\pm$ 0.84}} \\
  & Stress        &  81.50 &  81.81 &  81.65 & \\
  & Amusement     &  78.74 &  75.56 &  77.12 & \\
  \cmidrule(lr){2-5}
  & \emph{Macro Avg}     &  82.91 &  82.26 &  82.57 $\pm$ 1.30& \\
  & \emph{Weighted Avg}  &  84.75 &  84.80 &  84.77 $\pm$ 0.87& \\
  \midrule

\multirow{6}{*}{TCN}
  & Baseline      &  95.28 &  85.95 &  90.37 & \multirow{6}{*}{\centering{86.35 $\pm$ 0.53}} \\
  & Stress        &  81.64 &  90.97 &  86.05 & \\
  & Amusement     &  71.81 &  79.37 &  75.40 & \\
  \cmidrule(lr){2-5}
  & \emph{Macro Avg}     &  82.91 &  85.43 &  83.94 $\pm$ 0.71& \\
  & \emph{Weighted Avg}  &  87.23 &  86.35 &  86.56 $\pm$ 0.62& \\
\bottomrule
\end{tabular}
}
\label{tab:ablation_wrist_chest}
\vspace{-2.0ex}
\end{table}

\subsection{Ablation Studies on Sensor Modalities}

To quantify each modality's contribution, we selectively removed wrist- or chest-derived signals from the multimodal models. Comparing these unimodal variants against the full multimodal configuration shows the benefit of multimodal fusion and clarifies how each modality contributes to model performance and generalization across subjects.

\vspace{2.0ex}
\noindent \textbf{Component and Signal Importance}. To understand what drives model performance, we systematically evaluated the contributions of key architectural components, such as the fusion layer, attention mechanism, and the temporal encoder, as well as individual physiological signals by selectively ablating them from the full model. This analysis helps quantify the importance of each component and validate which design elements are critical to performance and which offer only marginal benefits for robust affect recognition.

\vspace{1.0ex}

\noindent \textbf{Modality Discrimination}. Although the ablation experiments isolate the technical contribution of each modality, we also analyze their physiological relevance. To assess the contribution of each sensor modality, we trained models using only wrist- or chest-based inputs. The wrist-worn sensors primarily capture \ac{EDA}, temperature, and motion through accelerometry, while the chest-mounted sensors measure respiration and \ac{ECG}. This contextual analysis helps interpret why combining both modalities enhances affect recognition.

\vspace{1.0ex}
\noindent \textbf{Practical Deployment}. Wrist-based devices, such as smartwatches, are more practical for real-world use because of their convenience and non-invasiveness. Demonstrating competitive performance using wrist-only input shows the feasibility of lightweight, wearable affect recognition systems.

\vspace{1.0ex}
\noindent \textbf{Findings}. As shown in Table~\ref{tab:ablation_wrist_chest}, our ablation results indicate that removing either sensor modality or disabling temporal modeling consistently leads to a drop in performance across all models, reinforcing the importance of multimodal fusion and temporal context. The results in this table are produced under the ablation protocol, in which wrist-only and chest-only inputs are reconstructed by slicing features from the fused multimodal tensor, and model configurations are intentionally simplified and standardized relative to the primary benchmark. Conclusions drawn here therefore reflect within-ablation comparisons and should not be interpreted as a direct replication of the primary unimodal results in Table~\ref{tab:unimodal_comparison}. The TCN model achieved the strongest wrist-only performance within this ablation setting (98.20 $\pm$ 0.41\% accuracy, 97.72 $\pm$ 0.52\% macro-F1 score), outperforming both LSTM and Transformer. LSTM also maintained high performance using wrist-only input (94.38 $\pm$ 0.88\% accuracy), with particularly high F1-scores for stress (97.81) and baseline (94.92), while amusement remained more difficult to classify. Across all models, amusement consistently resulted in the lowest scores, highlighting its subtle physiological expression and class imbalance. In chest-only configurations, performance declined further; for example, the Transformer model achieved 84.80 $\pm$ 0.84\% accuracy, suggesting that wrist signals may carry more discriminative information for affect recognition in this dataset. These findings indicate that while both sensor modalities contribute significantly, wrist signals alone can support accurate emotion recognition in constrained settings. TCN and LSTM architectures, in particular, remain effective even without multimodal input, reinforcing their suitability for real-time, wearable applications.

\subsection{Saliency-Based Interpretation}

Gradient-based saliency maps show that ACC and temperature  dominate wrist-only predictions, while the multimodal model assigns greater importance to RES, HR, and EDA under stress conditions. These observations are illustrative and subject-specific. Full saliency maps and population-level analysis are provided in the supplementary material.

\subsection{Summary}
\label{summary}

Our analysis shows several findings with practical and methodological implications. The ensemble achieves the highest overall performance (98.91 $\pm$ 0.13\% accuracy), with markedly lower inter-subject variance than individual models. Wrist-only configurations offer competitive performance in resource-constrained settings, while multimodal fusion consistently improves robustness. A detailed comparison with previous work on WESAD, including summary-statistics baselines and previous classical machine learning and deep learning results, is provided in the supplementary material, where our Transformer and ensemble models consistently outperform previously reported results under the same LOSO-CV protocol.

\section{Conclusion}
\label{conclusion}

We evaluated the effectiveness of various deep learning architectures and sensor modalities for automatic emotion recognition using physiological signals. We conducted a comprehensive comparison of three model architectures, LSTM, TCN, and Transformer, across unimodal (wrist-only and chest-only) and multimodal input settings using the WESAD dataset. Our results show that early fusion of chest and wrist signals improves performance over unimodal baselines, demonstrating the complementary nature of the two sensor types. We implemented a late-fusion ensemble strategy by averaging the predictions of the three architectures trained on fused inputs, which yielded the highest overall accuracy (98.91 $\pm$ 0.13\%) and macro-F1 score (98.56 $\pm$ 0.17\%). Through ablation studies and ensemble experiments, we analyzed how architectural and modality choices influence performance, generalization, and practical applicability. Our findings show that Transformer-based models achieved strong generalization under the LOSO-CV protocol and competitive performance in multimodal settings. Wrist-only models offer a practical trade-off suitable for lightweight, wearable applications. These results show the practical value of ensemble-based fusion in wearable affect recognition, particularly for systems designed to generalize across diverse users. Amusement remained the most difficult emotion to classify, likely due to overlapping physiological patterns and limited training samples.

\section{Limitations and Future Work}
\label{future_work}

\noindent \textbf{Limitations}. One main limitation of this study is the use of a single dataset collected under controlled laboratory conditions. Although WESAD enables a fair and controlled cross-architecture comparison, which is the primary focus of this work, future evaluations on datasets with greater diversity in subjects, devices, and real-world conditions are needed to further validate the generalizability of the proposed framework. This setting may not fully capture the variability, noise, and complexity of real-world emotional states and physiological sensor readings. Furthermore, class imbalance and potential variability across sensor recordings introduce additional challenges to generalization.

\vspace{1.0ex}

\noindent \textbf{Future Work}. Future work will focus on validating the proposed framework on more diverse datasets, including ambulatory and real-world settings, to better assess generalizability beyond controlled laboratory conditions. We will also explore multi-resolution representations that combine model predictions across temporal resolutions, preserving both low-frequency affective dynamics and high-frequency morphological features, as our sensitivity analysis shows that different architectures respond differently to resampling rate. Improving model interpretability through population-level techniques such as SHAP and Integrated Gradients remains an important direction to support clinical and behavioral applications.

\bibliographystyle{ACM-Reference-Format}
\bibliography{ACM-BCB-References}

\clearpage

\appendix

\renewcommand{\thesection}{ \Alph{section}}
\section{Supplementary Material}

This supplementary material provides additional experimental results, baseline comparisons, and interpretability analyses referenced in the main paper. All experiments follow the same LOSO-CV protocol and preprocessing pipeline described in Section 3 of the main paper.

\subsection{Sensitivity Analysis: Resampling Rate}
\label{sec:resampling}

To empirically assess the impact of the 4 Hz resampling choice on model performance, we evaluated all three architectures under a 16 Hz resampling rate with proportionally scaled windows of 40 time steps, maintaining the same 2.5-second temporal coverage as the primary 4 Hz setting. The results are summarized in Table~\ref{tab:resampling_comparison}. LSTM and TCN show marginal improvements at 16 Hz (98.30\% $\pm$ 0.26\% and 98.60\% $\pm$ 0.24\% accuracy respectively), while the Transformer performs slightly better at 4 Hz (99.02\% $\pm$ 0.51\% vs 97.21\% $\pm$ 0.64\%). The standard deviation values decrease at 16 Hz for LSTM and TCN, suggesting more 
consistent generalization across subjects at higher resolution. These architecture-specific differences are modest and fall within overlapping standard deviation ranges, indicating that 
performance is robust across resampling rates. The 4 Hz setting was 
therefore retained as the primary configuration to maintain 
temporal alignment across modalities, reduce computational cost, 
and ensure consistency across models. Our sensitivity analysis shows that different architectures respond differently to resampling rate, supporting multi-resolution modeling as an important direction for future work.

\begin{table}[ht]

\caption{Sensitivity analysis comparing multimodal 
classification performance at 4 Hz and 16 Hz resampling rates on the 
WESAD dataset. Windows of 10 and 40 time steps are used respectively, 
maintaining identical 2.5-second temporal coverage. All metrics are 
reported in percentage (\%).}

\resizebox{\columnwidth}{!}{
\begin{tabular}{l|c|cc|cc}
\toprule
\textbf{Model} & \textbf{Rate} & \textbf{Accuracy} & \textbf{Macro-F1} & 
\textbf{Weighted-F1} \\
\midrule
\multirow{2}{*}{LSTM} 
  & 4 Hz  & 97.70 $\pm$ 0.55 & 97.09 $\pm$ 0.72 & 97.67 $\pm$ 0.56 \\
  & 16 Hz & 98.30 $\pm$ 0.26 & 97.98 $\pm$ 0.35 & 98.30 $\pm$ 0.26 \\
\midrule
\multirow{2}{*}{Transformer}
  & 4 Hz  & 99.02 $\pm$ 0.51 & 98.88 $\pm$ 0.64 & 99.02 $\pm$ 0.51 \\
  & 16 Hz & 97.21 $\pm$ 0.64 & 96.61 $\pm$ 0.76 & 97.20 $\pm$ 0.65 \\
\midrule
\multirow{2}{*}{TCN}
  & 4 Hz  & 96.76 $\pm$ 0.59 & 96.10 $\pm$ 0.76 & 96.75 $\pm$ 0.60 \\
  & 16 Hz & 98.60 $\pm$ 0.24 & 98.23 $\pm$ 0.33 & 98.61 $\pm$ 0.24\\
\bottomrule
\end{tabular}
}
\label{tab:resampling_comparison}

\end{table}

\subsection{Summary-Statistics Baselines}
\label{sec:baselines}

To assess whether the observed performance gains of temporal deep learning models can be explained by simple signal-level  differences, we evaluated two summary-statistics-based baselines under the same preprocessing pipeline and LOSO-CV protocol used in our multimodal experiments. For each 10-step window, we extracted summary features including the mean, standard 
deviation, minimum, maximum, and first-to-last differences across all signal channels, resulting in a 24-dimensional feature vector per window.

A Logistic Regression model trained on these features achieved 59.68\% $\pm$ 16.86\% accuracy and 52.93\% $\pm$ 20.47\% macro-F1 across LOSO folds, indicating that linear relationships over aggregated features are insufficient to capture the discriminative structure of the data. A Random Forest classifier using the same feature representation achieved 67.50\% $\pm$ 10.89\% accuracy and 57.16\% $\pm$ 13.82\% macro-F1, showing modest improvement over the linear model but remaining substantially below all temporal deep learning models. Beyond the mean performance gap, both baselines show high inter-subject variance across LOSO folds. The accuracy of Logistic Regression shows a standard deviation of $\pm$16.86\% across subjects, and Random Forest $\pm$10.89\%, compared to $\pm$0.13\% for our ensemble. This indicates that summary-statistics representations fail to generalize consistently across individuals, whereas temporal architectures do. This suggests that temporal architectures learn generalizable affective representations rather than subject-specific signal patterns, which is the main empirical claim of our work.

These results confirm that the performance gains of our temporal models are not attributable to simple signal-level differences between affective conditions. Instead, they reflect the ability 
of temporal architectures to capture ordering, transition dynamics, and cross-modal dependencies that cannot be recovered from per-window summary statistics. Detailed classification metrics are reported in Table~\ref{tab:multimodal_performance_baselines}, and confusion matrices are provided in Table~\ref{table:confusion_matrices_baselines}. Mean $\pm$ standard deviation values are computed across the 15 LOSO folds, while per-class metrics are derived from pooled predictions across all test folds. These results are summarized in Section 4.5 of the main paper and provide empirical support for the contribution of temporal modeling.

\begin{table}[h]

\caption{Multimodal classification performance of 
summary-statistics baselines on the WESAD dataset. All metrics 
are reported in percentage (\%).}

\resizebox{\columnwidth}{!}{
\begin{tabular}{l|l|ccc|c}
\toprule
\textbf{Model} & \textbf{Class} & \textbf{Precision} & \textbf{Recall} & \textbf{F1-score} & \textbf{Accuracy} \\
\midrule

\multirow{6}{*}{Logistic Regression}
  & Baseline      & 65.61 & 72.86 & 69.05 & \multirow{6}{*}{59.68 $\pm$ 16.86} \\
  & Stress        & 55.82 & 51.06 & 53.33 & \\
  & Amusement     & 40.87 & 32.82 & 36.41 & \\
  \cmidrule(lr){2-5}
  & \emph{Macro Avg}     & 54.10 & 52.25 & 52.93 $\pm$ 20.47 & \\
  & \emph{Weighted Avg}  & 58.52 & 59.59 & 58.85 $\pm$ 19.90 & \\

\midrule

\multirow{6}{*}{Random Forest}
  & Baseline      & 73.95 & 77.82 & 75.84 & \multirow{6}{*}{67.50 $\pm$ 10.89} \\
  & Stress        & 62.34 & 76.41 & 68.66 & \\
  & Amusement     & 44.83 & 19.29 & 26.97 & \\
  \cmidrule(lr){2-5}
  & \emph{Macro Avg}     & 60.37 & 57.84 & 57.16 $\pm$ 13.82 & \\
  & \emph{Weighted Avg}  & 65.58 & 67.57 & 65.48 $\pm$ 11.84 & \\

\bottomrule
\end{tabular}
}
\label{tab:multimodal_performance_baselines}

\end{table}

\begin{table}[h]

\caption{Confusion matrices for summary-statistics baselines 
on the multimodal WESAD dataset.}

\resizebox{\columnwidth}{!}{
\begin{tabular}{l|l|ccc}
\toprule
\multicolumn{2}{c|}{} & \multicolumn{3}{c}{\textbf{Predicted}} \\
\cmidrule(lr){3-5}
\textbf{Model} & \textbf{Actual Class} & \textbf{Baseline} & \textbf{Stress} & \textbf{Amusement} \\
\midrule

\multicolumn{5}{c}{\textbf{Multimodal (LOSO)}} \\
\midrule

\multirow{3}{*}{\centering Logistic Regression}
  & Baseline   & 5135 & 1173 & 740 \\
  & Stress     & 1632 & 2032 & 316 \\
  & Amusement  & 1059 & 435 & 730 \\
\cmidrule(lr){1-5}

\multirow{3}{*}{\centering Random Forest}
  & Baseline   & 5485 & 1245 & 318 \\
  & Stress     & 729 & 3041 & 210 \\
  & Amusement  & 1203 & 592 & 429 \\

\bottomrule
\end{tabular}
}
\label{table:confusion_matrices_baselines}

\end{table}

\subsection{Confusion Matrices for Deep Learning Models}
\label{sec:confusion}

Table~\ref{tab:confusion_dl} presents the confusion matrices for all deep learning models across wrist-only, chest-only, multimodal, and ensemble fusion settings. Each row shows predicted label counts per actual class.

\begin{table}[h]
\caption{Confusion matrices for deep learning models across 
all input settings on the WESAD dataset.}
\label{tab:confusion_dl}
\resizebox{\columnwidth}{!}{
\begin{tabular}{l|l|ccc}
\toprule
\multicolumn{2}{c|}{} & \multicolumn{3}{c}{\textbf{Predicted}} \\
\cmidrule(lr){3-5}
\textbf{Model} & \textbf{Actual Class} & \textbf{Baseline} & 
\textbf{Stress} & \textbf{Amusement} \\
\midrule
\multicolumn{5}{c}{\textbf{Wrist-Only}} \\
\midrule
\multirow{3}{*}{LSTM}
  & Baseline  & 1317 & 13 & 79 \\
  & Stress    & 14   & 781 & 2 \\
  & Amusement & 35   & 6  & 405 \\
\midrule
\multirow{3}{*}{Transformer}
  & Baseline  & 1298 & 59 & 52 \\
  & Stress    & 34   & 763 & 0 \\
  & Amusement & 143  & 11 & 292 \\
\midrule
\multirow{3}{*}{TCN}
  & Baseline  & 1385 & 4  & 20 \\
  & Stress    & 4    & 793 & 1 \\
  & Amusement & 14   & 5  & 426 \\
\midrule
\multicolumn{5}{c}{\textbf{Chest-Only}} \\
\midrule
\multirow{3}{*}{LSTM}
  & Baseline  & 1260 & 77 & 72 \\
  & Stress    & 45   & 712 & 40 \\
  & Amusement & 44   & 42 & 360 \\
\midrule
\multirow{3}{*}{Transformer}
  & Baseline  & 1260 & 104 & 45 \\
  & Stress    & 99   & 652 & 46 \\
  & Amusement & 65   & 44  & 337 \\
\midrule
\multirow{3}{*}{TCN}
  & Baseline  & 1211 & 102 & 96 \\
  & Stress    & 29   & 725 & 43 \\
  & Amusement & 31   & 61  & 354 \\
\midrule
\multicolumn{5}{c}{\textbf{Multimodal}} \\
\midrule
\multirow{3}{*}{LSTM}
  & Baseline  & 1399 & 4  & 6 \\
  & Stress    & 7    & 790 & 1 \\
  & Amusement & 42   & 1  & 402 \\
\midrule
\multirow{3}{*}{Transformer}
  & Baseline  & 1394 & 8  & 7 \\
  & Stress    & 2    & 794 & 1 \\
  & Amusement & 8    & 0  & 438 \\
\midrule
\multirow{3}{*}{TCN}
  & Baseline  & 1368 & 17 & 24 \\
  & Stress    & 9    & 786 & 3 \\
  & Amusement & 27   & 6  & 412 \\
\midrule
\multicolumn{5}{c}{\textbf{Ensemble Fusion}} \\
\midrule
\multirow{3}{*}{Majority Voting}
  & Baseline  & 1400 & 1  & 8 \\
  & Stress    & 1    & 796 & 1 \\
  & Amusement & 15   & 3  & 427 \\
\bottomrule
\end{tabular}
}
\end{table}

\subsection{Comparison with Previous Work}
\label{sec:comparison}

Tables~\ref{tab:comparison_wesad_multimodal} and \ref{tab:comparison_wesad_unimodal} compare our models with previously reported results on WESAD under the same LOSO-CV protocol. Our Transformer and ensemble models consistently outperform all previous work across both multimodal and unimodal 
settings by a substantial margin. Minor methodological differences, such as window lengths and preprocessing steps, may affect the metrics reported across studies, but results are compared under the same LOSO-CV protocol to ensure consistency.

\begin{table*}[h]
\setlength{\tabcolsep}{2pt} 
\centering
\caption{Comparison with reported results on the WESAD dataset (multimodal, 3-class classification). All metrics are reported as originally stated.}

\label{tab:comparison_wesad_multimodal}
\begin{tabular}{lccccc}
\toprule
\textbf{Model} & \textbf{Setting} & $F_1$-score (\%) & Accuracy (\%) & Study & Cross-Validation Protocol\\
\midrule
Decision Tree & All modalities & $58.05\pm1.61$ & $63.56\pm1.73$ & WESAD (Schmidt et al.~\cite{schmidt2018introducing}) & LOSO-CV\\
Random Forest & All modalities & $64.08\pm1.68$ & $74.97\pm1.11$ & WESAD (Schmidt et al.~\cite{schmidt2018introducing}) & LOSO-CV\\
AdaBoost      & All modalities & $68.85\pm0.89$ & $79.57\pm0.93$ &  WESAD (Schmidt et al.~\cite{schmidt2018introducing}) & LOSO-CV\\
Linear Discriminant Analysis (LDA) & All modalities &71.56 & 75.80 &WESAD (Schmidt et al.~\cite{schmidt2018introducing}) & LOSO-CV\\
k-Nearest Neighbour (kNN)           & All modalities & 48.70 & 56.14 &  WESAD (Schmidt et al.~\cite{schmidt2018introducing}) & LOSO-CV\\

\midrule
Random Forest & All modalities  & 65.73 & 67.56 & Garg et al.~\cite{garg2021stress} & LOSO-CV \\
Support Vector Machine (SVM) & All modalities  & 59.64  & 59.56 & Garg et al.~\cite{garg2021stress} & LOSO-CV \\
k-Nearest Neighbour (kNN) & All modalities  & 58.14   & 65.00 & Garg et al.~\cite{garg2021stress} & LOSO-CV \\
Linear Discriminant Analysis (LDA) & All modalities  & 50.44  & 67.06 & Garg et al.~\cite{garg2021stress} & LOSO-CV \\
AdaBoost & All modalities  & 63.82  & 64.34 & Garg et al.~\cite{garg2021stress} & LOSO-CV \\

\midrule
\multicolumn{6}{c}{\textbf{Our approaches}} \\

\midrule
LSTM & All modalities & 97.09 $\pm$ 0.72& 97.70 $\pm$ 0.55& This work & LOSO-CV\\
Transformer & All modalities & 98.88 $\pm$ 0.64& 99.02 $\pm$ 0.51& This work & LOSO-CV\\
TCN & All modalities & 96.10 $\pm$ 0.76& 96.76 $\pm$ 0.59& This work & LOSO-CV\\
Ensemble Fusion & All modalities & 98.56 $\pm$ 0.17& 98.91 $\pm$ 0.13& This work & LOSO-CV\\
\bottomrule
\end{tabular}
\end{table*}


\begin{table*}[h]
\setlength{\tabcolsep}{2pt} 
\centering
\caption{Performance comparison of unimodal models on the WESAD Dataset. All metrics are reported as originally stated.}
\label{tab:comparison_wesad_unimodal}
\begin{tabular}{lccccc}
\toprule
\textbf{Model} & \textbf{Setting} & $F_1$-score (\%) & Accuracy (\%)& Study & Cross-Validation Protocol\\
\midrule
Decision Tree & All wrist & $43.62\pm1.33$ & $53.98\pm1.79$ & WESAD (Schmidt et al.~\cite{schmidt2018introducing}) & LOSO-CV\\
Random Forest & All wrist & $62.86\pm0.65$ & $74.85\pm0.20$ & WESAD (Schmidt et al.~\cite{schmidt2018introducing}) & LOSO-CV\\
AdaBoost      & All wrist & $64.12\pm0.98$ & $75.21\pm0.77$ &  WESAD (Schmidt et al.~\cite{schmidt2018introducing}) & LOSO-CV\\
Linear Discriminant Analysis (LDA) & All wrist & 63.24 & 70.74 & WESAD (Schmidt et al.~\cite{schmidt2018introducing}) & LOSO-CV\\
k-Nearest Neighbour (kNN)           & All wrist & 37.20 & 45.54 &  WESAD (Schmidt et al.~\cite{schmidt2018introducing}) & LOSO-CV\\

\midrule
Decision Tree & All chest & $53.06\pm0.50$ & $57.68\pm0.40$ & WESAD (Schmidt et al.~\cite{schmidt2018introducing}) & LOSO-CV\\
Random Forest & All chest & $60.80\pm1.00$ & $68.76\pm1.35$ & WESAD (Schmidt et al.~\cite{schmidt2018introducing}) & LOSO-CV\\
AdaBoost      & All chest & $64.89\pm0.81$ & $74.74\pm0.94$ &  WESAD (Schmidt et al.~\cite{schmidt2018introducing}) & LOSO-CV\\
Linear Discriminant Analysis (LDA) & All chest & 72.49 & 76.50 & WESAD (Schmidt et al.~\cite{schmidt2018introducing}) & LOSO-CV\\
k-Nearest Neighbour (kNN)           & All wrist & 38.39 & 46.18 &  WESAD (Schmidt et al.~\cite{schmidt2018introducing}) & LOSO-CV\\

\midrule
LDA & ECG-based  & 81.3 & 85.4 & Prajod et al.~\cite{prajod2022generalizability} & LOSO-CV \\
CNN (spectrogram)~\cite{liakopoulos2021cnn} & ECG-based  & 79.4 & 82.4 & Prajod et al.~\cite{prajod2022generalizability} & LOSO-CV \\
Transformer~\cite{behinaein2021transformer} (without fine-tuning) & ECG-based  & 69.7 & 80.4 & Prajod et al.~\cite{prajod2022generalizability} & LOSO-CV \\
Random Forest & ECG-based  & 81.3 & 86.3 & Prajod et al.~\cite{prajod2022generalizability} & LOSO-CV \\
SVM & ECG-based  & 83.2 & 87.1 & Prajod et al.~\cite{prajod2022generalizability} & LOSO-CV \\
Multi-Layer Perceptron (MLP) & ECG-based  & 85.9 & 89.5 & Prajod et al.~\cite{prajod2022generalizability} & LOSO-CV \\
ECG Emotion Model & ECG-based  & 85.8 & 89.7 & Prajod et al.~\cite{prajod2022generalizability} & LOSO-CV \\
Deep ECGNet & ECG-based  & 85.7 & 90.8 & Prajod et al.~\cite{prajod2022generalizability} & LOSO-CV \\

\midrule
\multicolumn{6}{c}{\textbf{Our Approaches}} \\

\midrule
LSTM & All wrist & 94.92 $\pm$ 0.53& 95.81 $\pm$ 0.48& This work & LOSO-CV\\
Transformer & All wrist & 91.87 $\pm$ 1.93& 93.33 $\pm$ 1.68& This work & LOSO-CV\\
TCN & All wrist & 89.43 $\pm$ 0.64& 91.06 $\pm$ 0.46& This work & LOSO-CV\\

\midrule
\multicolumn{6}{c}{\textbf{Ablation Study}} \\

\midrule

LSTM & All wrist & 93.21 $\pm$ 0.97 & 94.38 $\pm$ 0.88 & This work & LOSO-CV\\
Transformer & All wrist & 85.85 $\pm$ 2.46 & 88.73 $\pm$ 1.79 & This work & LOSO-CV\\
TCN & All wrist & 97.72 $\pm$ 0.52 & 98.20 $\pm$ 0.41& This work & LOSO-CV\\

\midrule
LSTM & All chest & 85.76 $\pm$ 1.04& 87.93 $\pm$ 0.70& This work & LOSO-CV\\
Transformer & All chest & 82.57 $\pm$ 1.30& 84.80 $\pm$ 0.84& This work & LOSO-CV\\
TCN & All chest & 83.94 $\pm$ 0.71& 86.35 $\pm$ 0.53& This work & LOSO-CV\\

\bottomrule
\end{tabular}
\end{table*}

\subsection{Saliency Maps}
\label{sec:saliency}

\begin{figure}[h]
\centering
\includegraphics[width=\columnwidth]{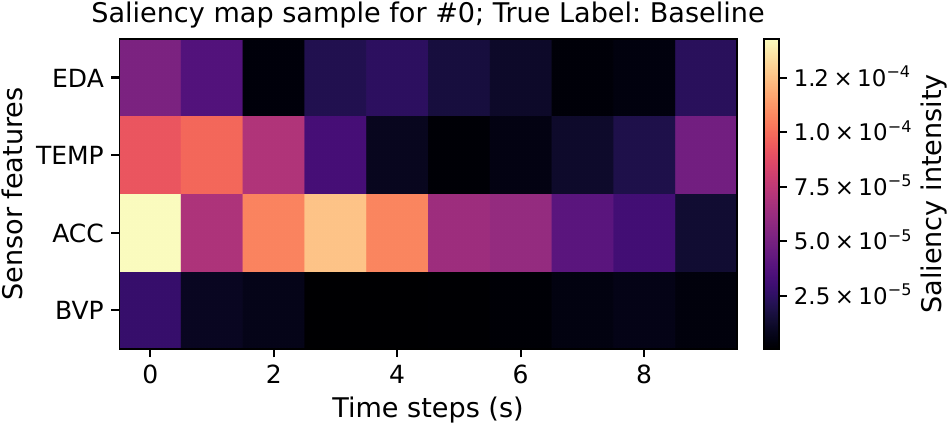}

\caption{Saliency map for a sample from the baseline class using the unimodal LSTM model.}

\label{saliency:unimodal_lstm}
\end{figure}

\begin{figure}[h]
\includegraphics[width=\columnwidth]{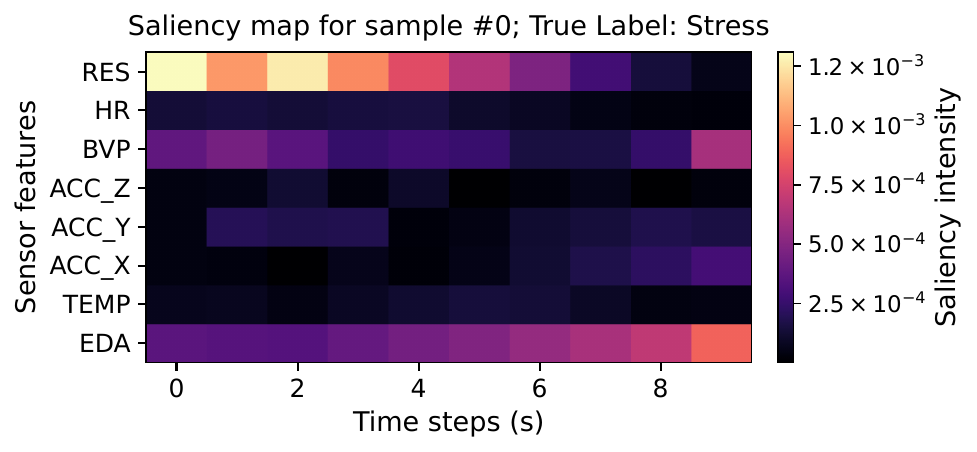}
\caption{Saliency map for a sample from the stress class using the multimodal LSTM model.}

\label{saliency:multimodal_lstm}
\end{figure}

Figures~\ref{saliency:unimodal_lstm} and 
\ref{saliency:multimodal_lstm} show gradient-based saliency maps for representative samples from the baseline and stress classes using the unimodal and multimodal LSTM models respectively. In the wrist-only setting, accelerometer (ACC) and temperature signals dominate the saliency maps, particularly for baseline detection. The multimodal model assigns greater importance to respiration (RES), heart rate (HR), and EDA, especially under stress conditions. These observations are illustrative and subject-specific, and should not be interpreted as 
population-level findings. A robust population-level interpretation would require techniques such as SHAP or Integrated Gradients across all subjects.

\end{document}